# Automated 3D Localization of the Femoral ACL Footprint in Knee MRI

Ruida Cheng[a], Gabriel Gibson[c], Ali Uneri[b], Frances T. Sheehan[c], Barry Boden[d]

[a] Scientific Application Services, Center for Information Technology, NIH, Bethesda, MD
[b] Department of Biomedical Engineering, Johns Hopkins University, Baltimore, MD
[c] Rehabilitation Medicine Department, Clinical Center, NIH, Bethesda, MD
[d] The Centers for Advanced Orthopaedics, Rockville, MD

## ABSTRACT

A major cause of anterior cruciate ligament (ACL) reconstruction failure is femoral tunnel malpositioning. Inaccurate ACL footprint localization and tunnel orientation may lead to meniscal pathology and osteoarthritis. Although deep learning methods have been applied to ACL segmentation and rupture classification based on pre- and post-operative magnetic resonance (MR) images, 3D footprint localization remains largely unexplored. To address this, we developed and compared two complementary approaches: a geometric model using a graph convolutional neural network applied to femoral meshes; and a volumetric landmark-detection model applied to 3D MR images. We evaluated the models on 7,969 knee images from a publicly available database. The image-based model outperformed the mesh-based model (average error 2.1 mm vs 2.9mm). For the volumetric model, 95.6% of left-knee and 92.8% of right-knee predictions were within 5 mm of the reference location. These results demonstrate the feasibility of automated 3D ACL footprint localization and establish a foundation for image-guided tunnel planning.

## 1. INTRODUCTION

Approximately 150,000-400,000 anterior cruciate ligament (ACL) reconstructions are performed annually in the US [1]. Despite improvements in knee stability with reconstruction, persistent high rates of post-operative joint instability, abnormal kinematics, and OA remain due to altered knee joint mechanics [3]. Early ACL reconstruction surgeries were commonly performed using a transtibial technique. Yet, this technique was criticized for causing improper femoral tunnel placement, one of the most common factors leading to ACL revision [4]. Advances in ACL footprint identification led to renewed interest in improving femoral tunnel localization during ACL reconstructions [5] and tibial-independent drilling techniques were developed that more accurately reproduced native femoral anatomy [6]. However, ACL footprint identification remains a manual process.

Recent advances in AI enable automated analysis of magnetic resonance (MR) images for ACL analysis. Existing methods have primarily focused on classifying ACL tears. Liu et al. [7] used a cascaded pipeline of convolutional classifiers to select sagittal MR slices containing the ACL, identify the ACL, and classify the ACL state. MRNet [8] for tri-plane knee MR image analysis was developed to classify ACL and meniscal tears. Qu et al. [9] proposed a coarse-to-fine approach to localize ACL rupture and classify rupture types. They used 3D U-Net For ACL segmentation and a 2D YOLO model to define rupture points and types. Liu et al. [10] proposed an ACLNet model that integrated convolutional features with transformer-based bone morphological point clouds to localize ACL ruptures. Automated ACL segmentation remains challenging due to the ACL's small size, low-contrast boundaries, and post-injury appearance variability. Lee et al. [11] proposed a graph-cut-based approach with a shape prior and label refinement, which requires initial semi-manual seeding of the ACL region. Flannery et al. [12] used a 2D U-Net for intact ACL segmentation. MGACA-Net [13] modified the DeepLabv3 architecture with atrous spatial pyramid pooling and squeeze-and-excitation blocks to segment the ACL. Most solutions explored U-Net derivatives for 2D or 2.5D segmentation, and none were applied to locating the ACL footprint.

Developing 3D deep learning models that directly and accurately identify the femoral ACL footprint from 3D MR images can help bridge the gap between automated image analysis and surgical planning by providing anatomical targets for more accurate tunnel placement during ACL reconstruction. Such automatic footprint-identification models could be combined with emerging computer navigation and arthroscopic computer vision systems to improve ACL tunnel placement.

## 2. METHODS

Localizing the femoral ACL footprint is a highly imbalanced 3D landmark-detection problem. The target occupies only a small fraction of the entire surface mesh or image volume. Direct pointwise prediction provides extremely sparse supervision and may be sensitive to small spatial variations. We instead formulate localization as heatmap predictions. For each case, the reference center is represented by a Gaussian distribution that assigns greater probability to locations near the landmark. The geometric model (§2.1) predicts this distribution over mesh vertices, whereas the volumetric model (§2.2) generates a voxel-wise probability heatmap directly from the MR image. During inference, each predicted heatmap is decoded into a 3D coordinate in the original patient space. This provides spatially continuous supervision and allows both models to learn the surrounding anatomical context rather than identify a single positive vertex or voxel.

## 2.1 Geometric ACL Footprint Localization from Femoral Surface Meshes

The geometric model (Fig. 1) represents each femur as a point cloud and predicts the ACL footprint center as a vertex-wise landmark probability distribution. Based on the dynamic graph convolutional neural network (DGCNN) [15], the pipeline consists of mesh preprocessing, dynamic graph feature extraction, and vertex-based probability map estimation.

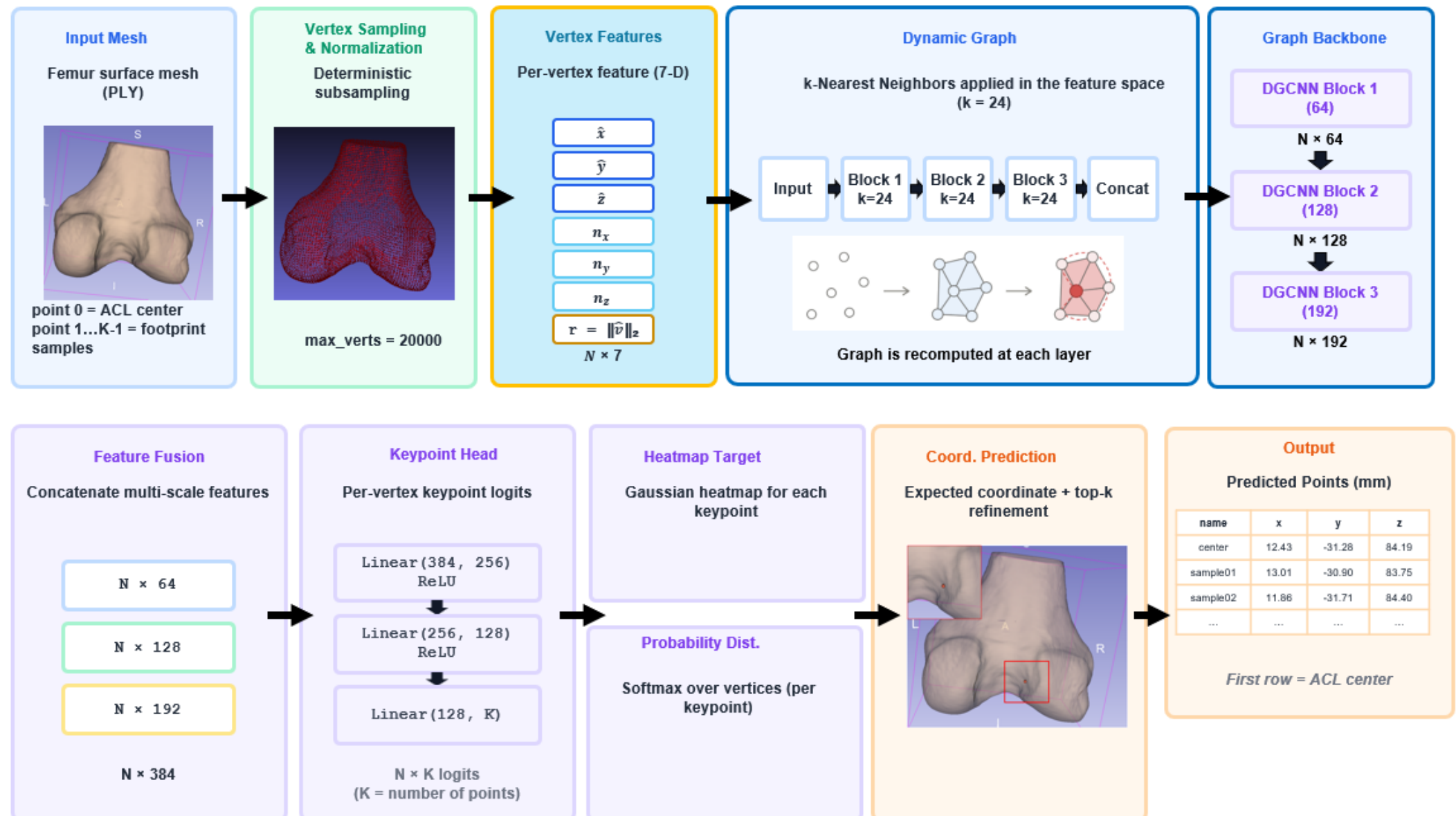

**Figure 1.** DGCNN for geometric ACL footprint center localization from femoral surface meshes.

**2.1.1 Mesh preprocessing.** For each knee, a femoral surface mesh in PLY format and a CSV file of annotated points are the model inputs. The landmark points consist of the ACL femoral footprint center and surrounding sampling points. The latter increases the likelihood of a spatially distributed center-point feature representation in the 3D mesh. The femoral bone model mesh is typically composed of hundreds of thousands of vertices, making mesh-based graph convolution prohibitively expensive. Thus, we perform a preprocessing step to reduce the vertices with deterministic sampling and normalization. The deterministic sampling ensures the uniform spacing of each point. Each vertex's anatomical position and local surface orientation is represented by a seven-dimensional feature vector $f_i = [\hat{x}_i, \hat{y}_i, \hat{z}_i, n_x, n_y, n_z, r_i]$, where $\hat{x}$ is the normalized coordinates, $n$ is the surface normal, and $r_i$ is the radial distance from the normalized mesh origin.

***2.1.2 Dynamic graph feature extraction.*** The DGCNN model performs ACL footprint center detection on a 3D point cloud via edge convolution blocks, which dynamically recomputes the k-nearest neighbors at each network layer. The dynamic computation captures local surface geometry. The shallow and deep layers connect vertices that lie close in physical distance and the learned feature space, respectively. The dynamic graph convolution neural network learns the multiscale geometric representations and generates a vertex-wise probability distribution.

***2.1.3 Vertex-wise probability estimation.*** The multiscale vertex features pass through the multilayer perceptron to generate one localization score (logits, Fig. 1) for each keypoint (center and sampled points) per vertex. This generates an $N \times K$ tensor, where $K$ columns represent the spatial likelihood map over the 3D femoral mesh. During training, each keypoint is represented by a soft Gaussian target, instead of a single positive vertex, where vertices closer to the center of the footprint have higher probabilities. The probabilities drop to nearly zero as the vertices move further from the center.

The DGCNN network extracts multi-scale geometric features and generates a vertex-wise probability distribution map for each target landmark point (center and sampling points). The training process combines Gaussian heatmap supervision with a coordinate regression loss, giving higher weights to points close to the ACL center. During inference, the predicted center coordinate is estimated from all sampling vertices, with refined estimate vertices derived from the highest probability map. Based on the original dynamic graph CNN model, we integrate the network with deterministic mesh sampling, mesh normalization, 7D vertex features, an increased receptive field in the DynamicEdgeConv backbone, and KL-divergence heatmap loss with coordinate regression loss. The overall mechanism ensures the geometrically and anatomically correct detection of the ACL footprint center on the 3D femur surface meshes.

### 2.2 Volumetric ACL Footprint Localization from 3D MR Images

The volumetric model (Fig. 2) operates directly on 3D MR images and predicts a voxel-wise probability heatmap for the ACL footprint center. We are motivated by the nnLandmark [16] architecture, which improves the localization of small and ambiguous targets in 3D medical images. We developed a framework that extends the original nnLandmark architecture with squeeze-and-excitation (SE) attention [17], a multi-scale high-resolution fusion module, and top-K peak-decoding building blocks to further improve detection of small and spatially ambiguous targets, such as the ACL footprint center in 3D MR images. With a new 3D MR image volume, the proposed architecture generates a heatmap channel per target landmark. After applying sigmoid activation, each voxel heatmap intensity represents the predicted probability.

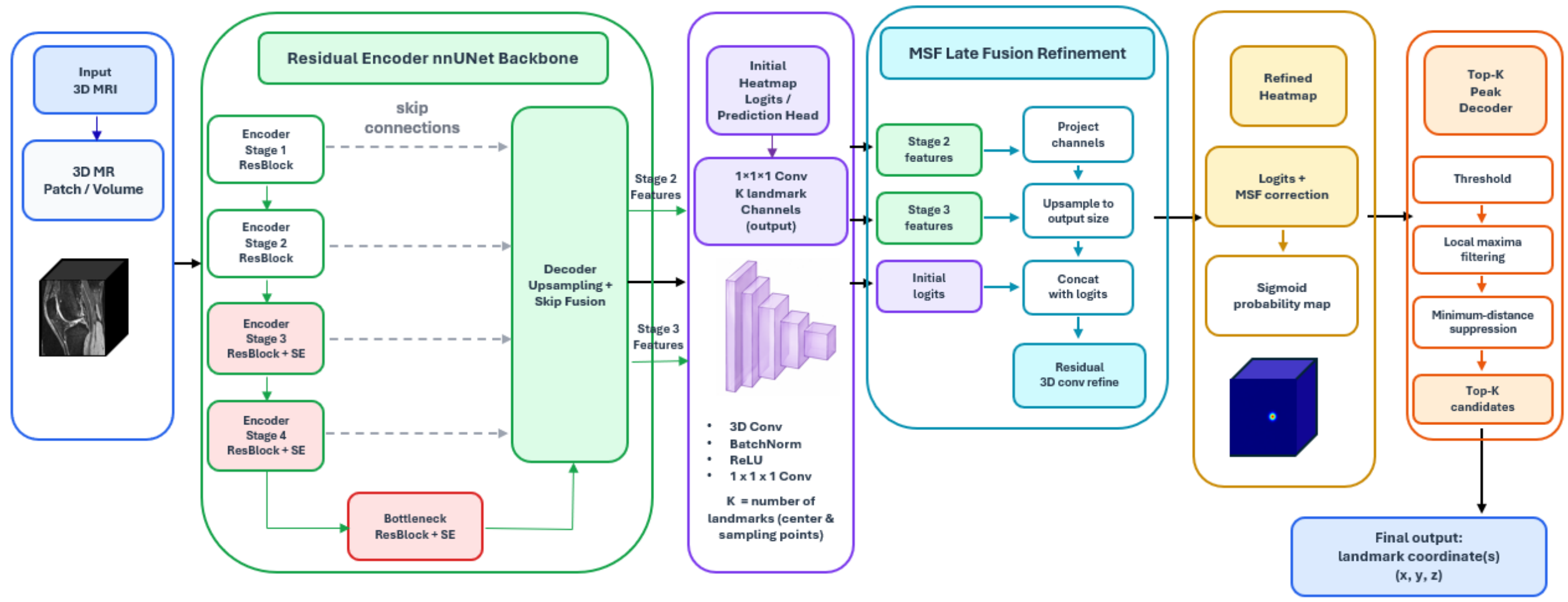

**Figure 2.** nnLandmark for volumetric ACL footprint center localization from 3D MR images.

***2.2.1 Backbone.*** The landmark detection architecture backbone is based on the conventional 3D residual-based nnU-Net [18]. The encoder's shallow layers capture local spatial features, whereas the deeper layers encode more abstract anatomical and contextual global features. The bottleneck extends with SE blocks to provide the largest receptive field and capture global context. The decoder progressively upsamples the bottleneck representations and merges them with the encoder-side feature maps via skip connections. The baseline backbone outputs an initial set of landmark heatmap logits from a 1×1×1 convolutional prediction head.

***2.2.2 Squeeze-and-excitation attention.*** To improve feature discrimination, Squeeze-and-excitation attention blocks are integrated into the deeper encoder stages and bottleneck. The SE blocks enable channel-wise attention by recalibrating the feature maps and capturing channel-wise dependencies. The squeeze operator applies global average pooling to generate the channel-wise local descriptors on the input feature maps. The excitation operator maps the input-specific descriptors to a set of channel weights to generate the final channel-wise attention. The SE blocks amplify more informative channel features and suppress noise and less relevant features for landmark detection.

***2.2.3: Multi-scale high-resolution fusion.*** The backbone decoder outputs the first heatmap logits via a 1×1×1 prediction-head convolution layer. Nevertheless, the decoder features may not possess the resolution needed to localize a small target, such as the ACL footprint center. Thus, a multi-scale fusion module is used to improve the first heatmap prediction. The intermediate features of the second and third encoders are taken and converted to the common feature size with 1×1×1 convolutions. These features are then resized to the final heatmap resolution with trilinear interpolation, and concatenated with the first heatmap logits. Further processing by the residual 3D convolution blocks leads to the final heatmap.

***2.2.4 Top-K peak decoding.*** Inference is done by converting the processed logits score into a probability heatmap using sigmoid activation followed by top-k peak selection decoding. Voxels are first filtered using the probability threshold ($\tau$ = 0.10). The spatial local maxima are then defined based on peak detection using a 3D max-pooling method with the neighborhood defined by the minimum-distance parameter. Ranked candidates are processed in descending order of score. Any candidate lying within the minimum distance ($r$ = 5 voxels) of an already-selected peak is suppressed to remove duplicate detections. The K highest-scoring peaks are retained (k = 5 by default); the top-ranked peak serves as the primary landmark prediction. The resulting point is then converted back to the original image space and reported in the patient/world coordinate systems.

The proposed nnLandmark-enhanced detection architecture consists of three integrated building blocks: SE attention highlights the importance of choosing channels in the deeper encoder and bottleneck; multi-scale fusion helps recover the spatial feature representation of the high resolution in the final heatmap; top-k decoding helps ensure more stable inference for the single-voxel candidate. The three building blocks are intended to improve the model's ability to detect small landmarks such as the ACL footprint center point.

### 2.3 Dataset and Reference Landmark Generation

From the 3DReasonKnee dataset [19], we acquired 7969 3D axial MR images (3D Double-Echo Steady-State acquisitions, 3T Siemens MAGNETOM Trio scanner, ~ 0.365×0.365×0.7 $mm^3$, 4,883 right & 3,087 left knees). The 3DReasonKnee dataset provides segmentation-based anatomical annotation masks, such as ACL and femur; however, it does not provide ACL femoral footprint center landmarks. Eighty percent of the images were used for model training and the remaining 20% for testing. We generated 3D femoral meshes and ACL footprint centers (as 3D landmark points) by developing a 3D mesh reconstruction and ACL center identification algorithms based on the ACL and femur segmentation masks. Each femoral mesh has one ACL footprint center and 20 surrounding sampling points, yielding 21 landmarks per case.

## 3. RESULTS

Both models produced clinical acceptable accuracies, however, the MR image-based outperformed the mesh-based model (mean error 2.09 vs 2.91mm, Table 1) when evaluated using the Euclidean distance (absolute mean error) between predicted and ground-truth landmark points. The same held true for the median error, 95th-percentile error (P95), and percentage of correct keypoints (PCK). Although some difficult cases showed errors of more than 20 mm, their frequency was low and had little impact on overall performance.

| Metric | Geometric Model (DGCNN) Left | Geometric Model (DGCNN) Right | Volumetric Model (nnLandmark) Left | Volumetric Model (nnLandmark) Right |
|---|---|---|---|---|
| Mean Error (mm) | 3.17 | 2.64 | 1.82 | 2.35 |
| Median Error (mm) | 2.38 | 2.01 | 1.28 | 1.60 |
| P95 Error (mm) | 6.93 | 6.20 | 4.60 | 6.82 |
| PCK @ 2 mm | 44.8% | 47.7% | 77.8% | 68.7% |
| PCK @ 3 mm | 58.6% | 72.6% | 89.9% | 83.5% |
| PCK @ 5 mm | 89.7% | 90.0% | 95.6% | 92.8% |

**Table 1.** 3D geometric and volumetric ACL footprint landmark detection performance.

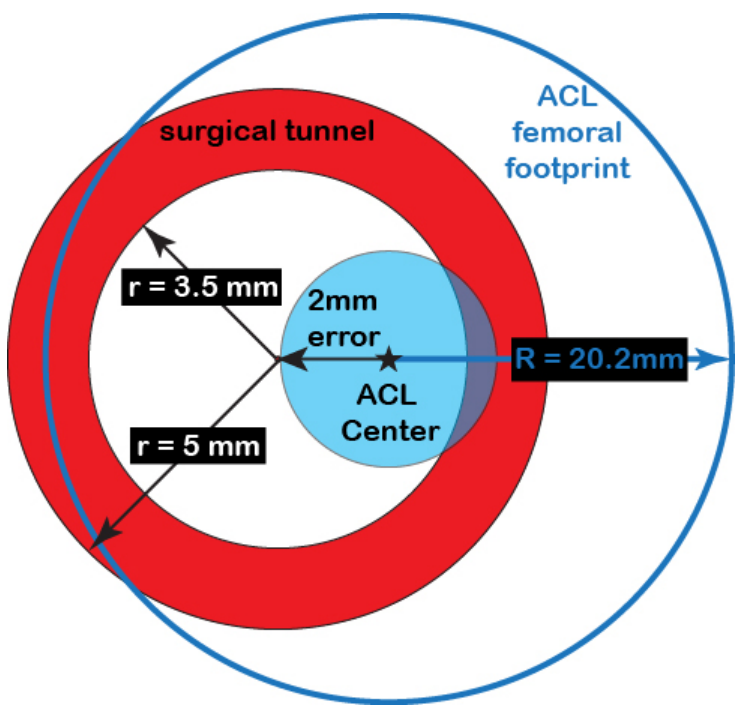

**Figure 3.** ***Idealized ACL femoral footprint schematic*** with drill hole placed at ACL footprint center with an added error. The ACL footprint is assumed circular (blue unfilled circle) with an area of 126.8 $mm^2$ (R=20.2mm) [20]. Typical ACL drills holes range for 7mm (white circle within red) to 10 mm (red circle). Then assuming an error (d = 2 mm, blue circle), the amount of overlap between the footprint and drill hole can be calculated as:

$$A_{\cap} = \pi r^2 - \left[ r^2 \cos^{-1}\left(\frac{d^2 + r^2 - R^2}{2dr}\right) + R^2 \cos^{-1}\left(\frac{d^2 - r^2 + R^2}{2dR}\right) - \frac{1}{2}\sqrt{(-d + r + R) * (d + r - R) * (d - r + R) * (d + r + R)} \right]$$

## 4. CONCLUSIONS

This pilot study illustrated the feasibility of applying 3D deep learning for automatic ACL femoral footprint center detection using two complementary models. Of the models, the 3D volumetric image-based nnLandmark detection was superior to the 3D mesh geometric DGCNN. Assuming a perfectly round femoral footprint of 126.8 $mm^2$ [20] (Fig. 3), our image- and mesh-based models' average errors keep a 7 mm drill hole within the femoral footprint. If the drill size is increased to 10 mm, then just 6% and 14% of the hole falls outside of the ACL footprint for the image- and mesh-based model predictions. These minor deviations, indicate that the prediction errors are within clinical acceptable limits. The mesh-based model, although slightly less accurate, holds significant promise in locating the femoral ACL footprint in individuals with a torn ACL, where the original footprint may not be visible on the images. We are investigating the generalizability of the image-based deep learning model to handle torn ACLs, as well as enhancements to further reduce errors. Therefore, this study has shown that 3D deep learning provides a feasible clinical approach to aid ACL footprint localization, improving drill site selection accuracy for ACL reconstruction footprint.

# ACKNOWLEDGMENTS

We sincerely thank Dr. Alexandra Ertl and Dr. Fabian Isensee for providing the nnLandmark training implementation code that supported this study.

## APPENDIX

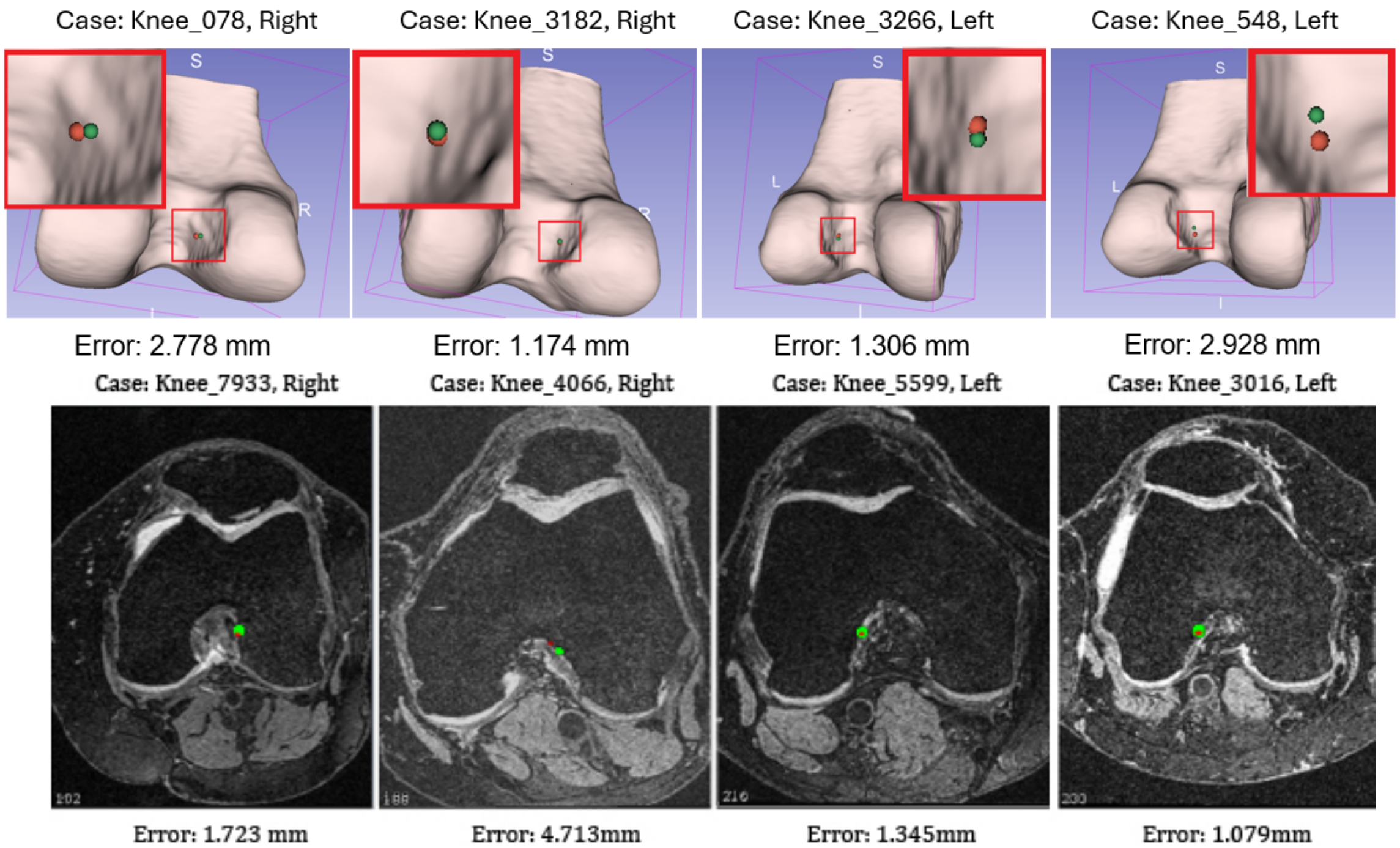

**Figure 1.** ***3D Image-based & Mesh-based ACL footprint landmark detection results***. The green sphere represents the ground truth ACL center. The red color sphere is the prediction. The error between the two is listed below each model